\documentclass[11pt,a4paper]{article}
\usepackage[hyperref]{acl2020}
\usepackage{times}
\usepackage{latexsym}

\usepackage{microtype}

\usepackage{subcaption}
\usepackage{latexsym}
\usepackage{amsmath}
\usepackage{multirow}
\usepackage{tablefootnote}
\usepackage{comment}
\usepackage{amssymb}
\usepackage{url}
\usepackage{enumitem}
\usepackage{array}
\usepackage{diagbox}
\usepackage{tikz}
\setlist{nosep}

\usepackage{color}

\usepackage{graphicx,calc}
\DeclareMathOperator*{\argmax}{arg\,max}

\aclfinalcopy 
\def\aclpaperid{1255} 

\title{Zero-Shot Learning of Text Adventure Games with Sentence-Level Semantics}

\author{Xusen Yin \\
ISI/USC \\
4676 Admiralty Way \\
Marina del Rey, CA 90292 \\ 
  \texttt{xusenyin@usc.edu} \\\And
  Jonathan May \\
  ISI/USC \\
 4676 Admiralty Way \\
Marina del Rey, CA 90292 \\ 
  \texttt{jonmay@isi.edu} \\}
  
\date{}

\begin{document}
\maketitle
\begin{abstract}

Reinforcement learning algorithms such as Q-learning have shown great promise in training models to learn the optimal action to take for a given system state; a goal in applications with an exploratory or adversarial nature such as task-oriented dialogues or games. However, models that do not have direct access to their state are harder to train; when the only state access is via the medium of language, this can be particularly pronounced. We introduce a new model amenable to deep Q-learning that incorporates a Siamese neural network architecture and a novel refactoring of the Q-value function in order to better represent system state given its approximation over a language channel. We evaluate the model in the context of zero-shot text-based adventure game learning. Extrinsically, our model reaches the baseline's convergence performance point needing only 15\% of its iterations, reaches a convergence performance point 15\% higher than the baseline's, and is able to play unseen, unrelated games with no fine-tuning. We probe our new model's representation space to determine that intrinsically, this is due to the appropriate clustering of different linguistic mediation into the same state.
\end{abstract}

\section{Introduction}
\label{sec:intro}

Reinforcement learning (RL) learns policies that tell agents to pick a right move at each game state to solve games. In particular Q-learning, the policy is to choose the action which will lead to the maximum total reward given a game state and all available actions.  Two key steps of using Q-learning to play games are game state understanding and action selection. Some games, e.g. Chess and Go, have perfect state information, since their states are all explicitly presented, but the enormous number of states make them difficult to explore thoroughly. Other games, like action or text adventure video games, have obscured game states, because the signal conveying the state (pixels on a screen or a continuous stream of text) is only an approximation of the true states. In both of these cases the role of \textit{function approximation}  converts an inappropriate state signal (whether from sparseness or noisiness) into a state representation proxy that is amenable to adequate action prediction. \textit{Deep Q-learning} \cite{mnih-atari-2013} primarily differs from classical Q-learning in that function approximation is accomplished by a multi-layer network, e.g. a Convolutional Neural Network (CNN) over image frames for Atari games \cite{mnih-atari-2013} and, for text adventure games, a CNN \cite{DBLP:conf/nips/ZahavyHMMM18,Yin2019ComprehensibleCT} or Long Short-Term Memory LSTM \cite{lstm-original} over the sequences of text provided by game state and player input \cite{D15-1001,DBLP:journals/corr/abs-1806-11525}.\footnote{Recent deep Q-learning approaches for perfect knowledge games use CNNs over image input or game board representations  \cite{silver2016mastering,Silver1140}, which we mention only for completeness, as our focus in this work is on encoding noisy sources, specifically English language sources.} While end-to-end optimization of these representations with feedback from game rewards over many experience replays can enable them to model game state sufficiently to play the game they are trained on well, no \textit{common ground} is learned by training, so that the models cannot easily transfer their learning between games of different genres. 


In this work we constrain our investigation to that of function approximation for the imperfect natural language signal, which is used in task-oriented dialogues and text adventure games. We specifically focus on text adventure games, which use pure texture interface to interact between a player and a game instance, in this paper. Text adventure games provide game information, e.g. observations of environment, physical status of a player, total scores earned, and response to the player's actions (mostly text-based commands). For normal text-based games, a player can only observe the dialogue history between the game and the player of a game playing (which we call a \textit{trajectory}) and scores at each step. A player must approximate game state from only this information to make decisions. 


The problem is that the game dialogues are varied with different actions to choose, length of each dialogue, and possible stochastic events. As a result, the state-language mapping is usually many-to-many, hampering robust state approximation. As a trivial example, a player that walks repeatedly eastward into a wall may generate trajectories consisting of the repeated sequence  [{\it go east} (from the player), {\tt you cannot go that way} (from the game)]. Each trajectory corresponding to $i$ repetitions of this sequence is different, but the game state does not change. In a task order-irrelevant game, the game state should also be insensitive to the order of the sub-trajectories generated in the completion of each task, but different order will yield a different trajectory.

We introduce a new model that mitigates the problem by explicitly learning a text-to-state transformation; we do this with the Siamese Neural Network (SNN) \cite{bromley1994signature}, which was previously applied in the context of object recognition in images \cite{Koch2015SiameseNN} and used for data augmentation when the label space is large but training data scarce for some classes. The approach 
is helpful in one-shot evaluations, as it learns the general and easily transferable property of abstracting away unimportant details from signal. Our novel model integrates SNNs into a Deep Q-learning framework which simultaneously learns the policy that solves games and robust language features that discriminate states from dialogues. The SNN uses the same text encoder as the RL algorithm, taking a pair of game dialogues as input, then feeds the difference of two encoded states into a nonlinear binary classifier to tell if the input pair of dialogues represent identical states or not (Section \ref{sec:dsqn}). It can be used to train with games for which some state information is possible, then, due to the SNN objective's ability to distinguish generally pertinent information from language signal only, applied to games where state information is no longer available. To accommodate the multi-task training nature of the algorithm, our model also contains a novel \textit{factorization} of the policy calculation which yields faster, higher quality, and models that transfer more readily.
 

To recap, the novel contribution of our paper are:
    
\begin{itemize}
    \item a flexible multitask RL model that integrates deep Q-learning and SNNs for text game learning
    \item a factorization of the Q-value policy function to distinguish between rewards that have different levels of time-sensitivity, which is crucial for good behavior in our multitask environment  
    \item validation of our results on unseen text games in the same genre as our training data as well as on games from a different genre that demonstrates our model's superior performance, faster convergence, and better transferability.
\end{itemize}

\section{Model}

Our model primarily consists of the integration of a Deep Q-Network, which is a common tool for learning to play a variety of games, including text games, with an SNN. \cite{Song2018FeatureLA} uses SNN on task description to predict RL agents transferring ability. They assume that a set of already well-trained RL agents exist and create binary labels of transferable or non-transferable by using these RL agents on different tasks, while our work create a co-training framework of DQN and SNN.  A high level overview of the architecture is shown in Figure~\ref{fig:archi-dsqn}. We describe each component, then describe how they may be integrated.

\subsection{Deep Q-Network for Text Games}

The influential Deep Q-Network (DQN) approach of learning simple action video games pioneered by \newcite{mnih-atari-2013} has motivated research into the limits of this technique when applied to other kinds of games. We follow recent work that ports this approach to \textit{text-based games} \cite{D15-1001,P16-1153,fulda2017affordance,DBLP:conf/nips/ZahavyHMMM18,DBLP:journals/corr/abs-1805-07274,DBLP:conf/cig/KostkaKKR17,DBLP:journals/corr/abs-1806-11525,DBLP:journals/corr/abs-1812-01628,Yin2019ComprehensibleCT}. The core approach of DQN as described by \newcite{mnih-atari-2013} is to build a \textit{replay memory} of partial games with associated scores, and to use this to learn a function $f_{DQN}: (S, A) \to \mathcal{R}$, where $f_{DQN}(s,a)$ is the Q-value\footnote{an estimate of the total game reward obtainable by taking action $a$ at state $s$ and continuing optimally thereafter \cite{sutton2018reinforcement}} obtained by choosing action $a \in A$ when in state $s \in S$; from $s$, choosing $\argmax_{a \in A} f_{DQN}(s,a)$ affords the optimal action policy. Typically the $\argmax$ action selection is also used at inference time, though  variants include sampling from the Q-value distribution \cite{P16-1153}, choosing random actions with some small probability 
\cite{mnih-atari-2013,DBLP:conf/nips/ZahavyHMMM18,Yin2019ComprehensibleCT}, or applying bandit methods \cite{yin2019learn}.

The function $f_{DQN}$ requires first encoding trajectories $t$ into hidden states $f_e(t)$ with a encoder $f_e(\cdot)$, then either projecting hidden states into a list of all possible actions \cite{mnih-atari-2013,DBLP:conf/nips/ZahavyHMMM18,Yin2019ComprehensibleCT}, or encodeing actions with another encoder and computing relevance with other hidden states \cite{P16-1153,yin2019learn}. So that we can apply our agent in different games with different sets of actions, and we follow the second method; we encode an action $a$ into a vector $f_a(a)$ with an LSTM, then compute Q-values by calculating a weighted relevance between $f_e(t)$ and $f_a(a)$: $f_{DQN}=f_e(t)W_{dqn}f_a(a)$, where $W_{dqn}$ is a trainable DQN parameter matrix. While there are other approaches that could be taken to generate actions or reduce the feasible action space, our focus in this work is on state approximation, so the action choice is held static.

\newcite{mnih-atari-2013} did not have access to the true state of their video games, so they used four consecutive frames as a proxy input. For our domain games we use the current trajectory (the dialogue history between player and game). For function approximation, the fixed-size representation of game state from sensor input, most related text-games work uses LSTMs \cite{D15-1001,DBLP:journals/corr/abs-1812-01628,DBLP:journals/corr/abs-1806-11525,DBLP:conf/cig/KostkaKKR17,DBLP:journals/corr/abs-1805-07274}, while \newcite{DBLP:conf/nips/ZahavyHMMM18} and \newcite{Yin2019ComprehensibleCT,yin2019learn} use CNNs to achieve greater speed in training.
We use the Transformer \cite{NIPS2017_7181} model in our paper, since this model is faster to train than the LSTM and has recently been shown to have superior performance in a number of NLU tasks.

The DQN is trained in an exploration-exploitation method ($\epsilon$-greedy) \cite{mnih-atari-2013}: with probability $\epsilon$, the agent chooses a random action (explores), and otherwise the agent chooses the action that maximizes the DQN function. The hyperparameter $\epsilon$ usually decays from 1 to 0 during the training process. The DQN agent collects partial play samples into the replay memory. The training process then samples a batch from the replay memory to train the DQN.

Since there are no true Q-values, DQN training is done in an iterative manner, with the expected Q-value estimated as 
\[Q(s, a) = r + \lambda \max_{a'} f_{DQN}(s', a'),\]

\noindent where $s'$ is the next state of $s$ applying the action $a$. In practice, we use a delayed $f_{DQN}$ \cite{Hasselt:2016:DRL:3016100.3016191} to compute the maximum Q-value from the next state $s'$. The predicted Q-value is $\hat{Q}(s, a)=f_{DQN}(s, a)$. We use square error loss to learn the DQN:
\[l_{DQN} = \|\hat{Q}(s, a) - Q(s, a)\|^2\]

\subsection{Siamese Neural Network}
We use an SNN \cite{Koch2015SiameseNN} to learn whether two trajectories are semantically equivalent or not:
\[f_{snn}(t_i, t_j) = \sigma \left(W_{snn} \left|f_e (t_i) - f_e (t_j)\right| + b\right)\]
\noindent where $f_e$ is a text encoder, $W_{snn}$ is a trainable parameter matrix, $b$ is the bias term, and $\sigma$ is the nonlinear sigmoid function. SNNs were introduced in a classification context especially for the scenario where $k$, the label space, may be large,  but there are fewer examples per class than desired. The SNN solves this problem by converting a $k$-wise classification problem over $n$ training examples into a binary classification problem over $O(n^2)$ training examples.

In our case,
we use the SNN in a multitask framework. The SNN task is to classify two trajectories $t_i$ and $t_j$ as semantically equivalent or not. This classifier is trained using 
sigmoid cross entropy loss $l_{snn}=-\log f_{snn}$ over a corpus of equivalence-labeled trajectory pairs.


We collect training trajectories as a byproduct of the DQN training from the replay memory. To determine state equivalence we use a hash of the current scene description and player inventory as a proxy; while still an approximation of state, this approach, which is inspired by \newcite{fulda2017affordance}, is a reasonable proxy for state equivalence. It has the property that, unlike trajectories, it \textit{overestimates} equivalence; for example, two states that differ by the location of an object that is out of the range of player's observation will be regarded as equivalent.
 During SNN training, we sample from the label space uniformly with replacement, then choose a trajectory tagged with the labels in that sample, to avoid label bias that would result from sampling directly from the experience replay trajectory pool. 
 For each trajectory that we sample, we sample another trajectory that has the same state label and a trajectory with a different state label, yielding a balanced SNN training corpus.

\section{DSQN: Multi-task Learning of DQN and SNN}
\label{sec:dsqn}

\begin{figure}
\centering
    \includegraphics[width=0.48\textwidth]{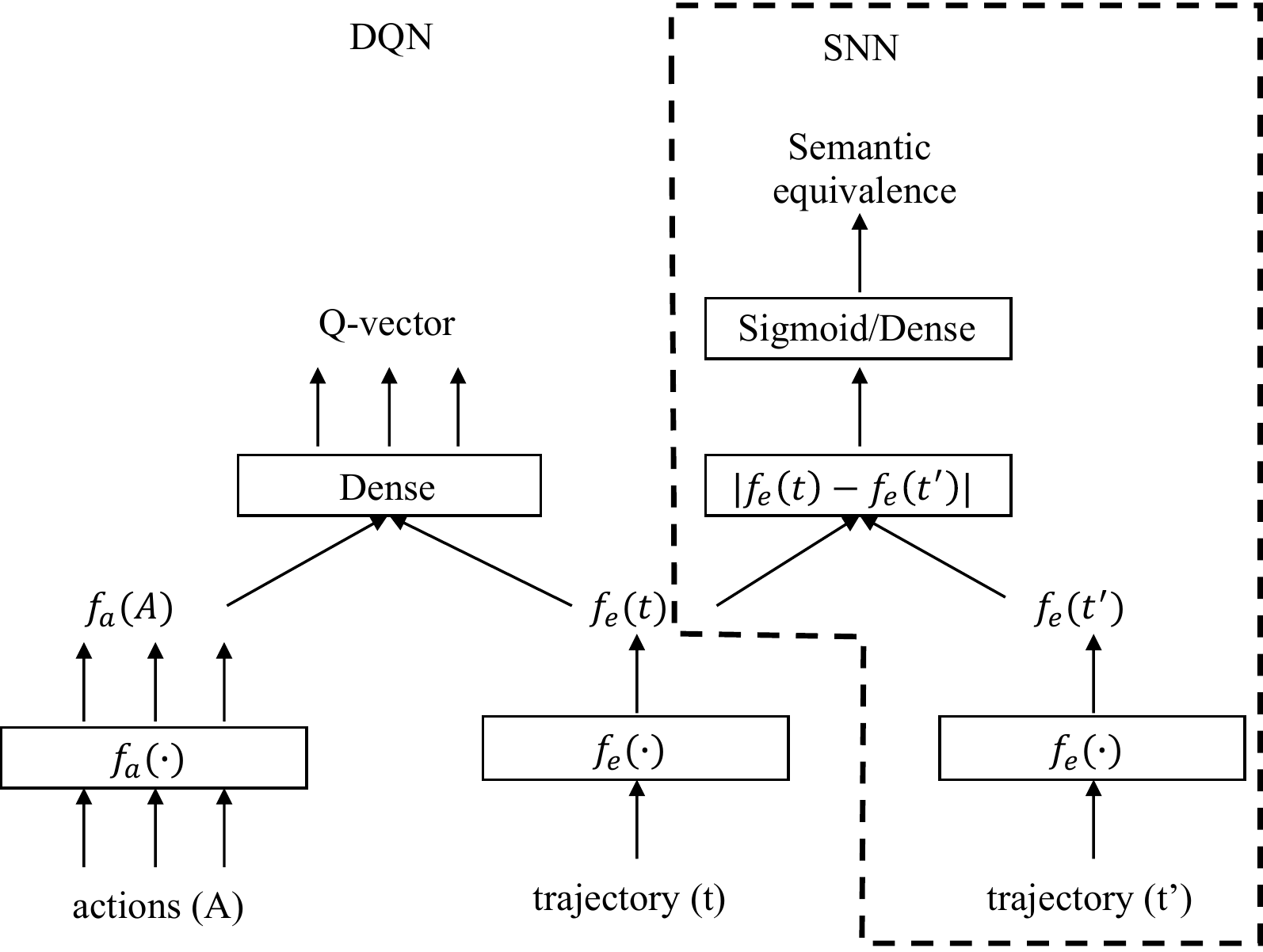}
    \caption{Multi-task learning framework for DSQN. The portion on the left depicts a normal DQN framework, while the boxed portion on the right is the SNN extension. The DQN takes a trajectory and actions as inputs and estimates Q-values of all these actions as output; this is trained by square error loss with iterative training. The SNN consumes two trajectories and makes a binary semantic equivalence prediction; this is optimized with sigmoid cross entropy loss. }
   \label{fig:archi-dsqn}
\end{figure}
In our multi-task learning framework, called DSQN, the SNN and DQN share the state encoder $f_e$, as shown in Figure \ref{fig:archi-dsqn}.   With the multi-task learning framework, we get two outputs at once during training, 1) the semantic equivalence of two trajectories, and 2) the Q-values for a trajectory and actions. Given the training of SNN and DQN, we have two loss functions to consider.

We follow \newcite{Kendall2017MultitaskLU} to organize the loss of DQN, which is a regression loss, and the loss of SNN, which is a classification loss, into the following form:
\[l =\frac{1}{2} e^{-s1} l_{dqn} + e^{-s2} l_{snn} + \frac{1}{2}s1 + \frac{1}{2}s2,\]
\noindent in which $s1$ and $s2$ are two trainable variables to control the weights between DQN and SNN losses.

This multi-task learning framework is rather flexible. During training we can choose to train the multi-task loss or we can choose to independently train each tasks. During evaluation, we can use both the SNN and DQN together (if state information is available) and we can also use only the DQN, if we just have access to trajectories.

\subsection{Q-value Factorization}
\label{sec:q-factorization}
\begin{figure}
\centering
    \includegraphics[width=0.48\textwidth]{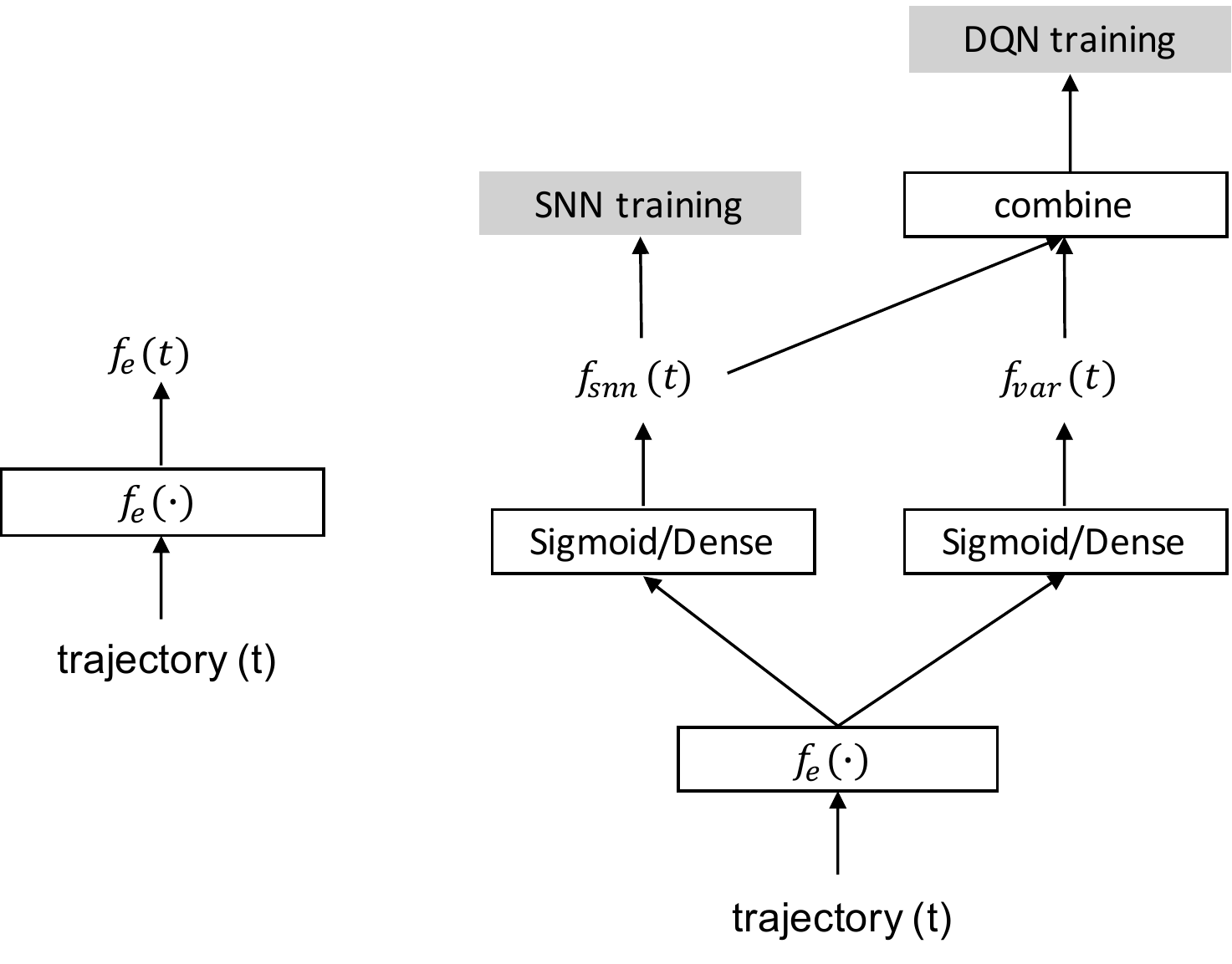}
    \caption{Q-factorization. After state encoder, the hidden state is further encoded with two different nonlinear dense layers, resulting in the $f_{snn}(t)$ for SNN, and another DQN hidden state $f_{var}(t)$ to allow time-sensitive rewards. The combination of the two hidden states is used as the hidden state vector for DQN training, while only the $f_{snn}(t)$ is used for SNN training.}
   \label{fig:q-factorization}
\end{figure}
The SNN training loss encourages $\sigma(W_{snn}|f_e(t1)-f_e(t2)|+b) = 0$ if trajectories $t1$ and $t2$ are semantically equivalent.
According to DQN, we have $Q(t, a) =  f_e(t) W_{dqn} f_a(a)$. So if t1 and t2 are semantically equivalent, then we have $f_e(t1) = f_e(t2)$ according to SNN, and consequently we have $Q(t1, a) = Q(t2, a)$.

In practical terms this means that non-essential exploration does not change game state. Consider a player in a room containing a book deciding to pick the book up; in trajectory terms this might be expressed as [\texttt{the room has a book}, \textit{pick up the book}, \texttt{you now have a book}]. If the player then looks at the book but this does not cause a state change, the trajectory is extended with, e.g. [\textit{look at the book}, \texttt{it has a red cover}]. From the game's perspective, the state has not changed.

However, this is in practice often in contradiction to alterations of the reward scheme used in game-solving RL models to encourage continuous reward feedback. These include \textit{discovery bonuses} that reward exploration \cite{DBLP:journals/corr/abs-1806-11525,NIPS2017_6868} and \textit{step penalties} that penalize repetition \cite{Yin2019ComprehensibleCT}. Incorporating such micro-state-change alterations along with SNN learning can either cause the SNN to be starved of useful training data (if each unique trajectory is regarded as a unique state) or to learn an incoherent representation (if the SNN and DQN disagree on state representation)

To accommodate the desires to use both SNN and reward alterations, we introduce two different nonlinear layers to reconstruct the encoded trajectories $f_e(t)$ into two states $f_{snn}(t)$ and $f_{var}(t)$, then add the separate Q-values together as the final Q-value.
We factorize the Q-value into the following form

\begin{equation*}
\begin{split}
Q(t, a) & = (f_{snn}(t) + f_{var}(t))W_{dqn}f_a(a) \\
        & = f_{snn}(t)W_{dqn}f_a(a) + f_{var}(t)W_{dqn}f_a(a) \\
        & = Q_{snn}(t, a) + Q_{var}(t, a) \\
\end{split}
\end{equation*}

\noindent and the SNN encoding is turned into
\[f_{snn}(t_i, t_j) = \sigma \left(W_{snn} \left|f_{snn} (t_i) - f_{snn} (t_j)\right| + b\right)\]
\noindent as shown in Figure \ref{fig:q-factorization}.
Our Q-factorization method is different from the Dueling DQN \cite{pmlr-v48-wangf16} that factorizes the value into a component that applies per state for all actions and another, state-action-specific component. Our factorization, by contrast, factors the value into two components that are state-action-specific, but that consider a decomposition of the state itself.







\section{Experiments}

We use the games released by Microsoft for the `First TextWorld Problems'\footnote{\url{https://www.microsoft.com/en-us/research/project/textworld}} competition. The competition provides 4,440 cooking games generated by the TextWorld framework \cite{textworld-a-learning-environment-for-text-based-games}. The goal of each game is to prepare a recipe. The action space is simple, yet expressive, and has a fairly large, though domain-limited, vocabulary. 

The games are divided into 222 different \textit{types}, with 20 games per type. A type is a set of attributes that increase the complexity of a game. These attributes include the number of ingredients, the set of necessary actions, and the number of rooms in the environment. 
A constant reward is given for each acquisition or proper preparation of a necessary ingredient as well as for accomplishing the goal (preparing the correct recipe), by each game. Each game has a different maximum score, so we report aggregate scores as a percentage of achievable points. The difficulty levels of each game vary according to different levels of cooking, manoeuvring, and ingredients acquirement skills required.

\subsection{Game Settings}

We hold out a selection of 10\% of the games and divide this portion into two separate test sets, each consisting of 222 games, one from each type. We randomly select an additional 400 games as a dev set and keep the remaining games for training. We consider an \textit{episode} to be a play-through of a game; there are multiple episodes of each game run during training and scores are taken over a 10-episode run of each game when evaluating test. An episode runs until a loss (an ingredient is damaged or the maximum of 100 steps is reached) or a win, by completing the recipe successfully. Apart from the inherent game reward, we add $-0.1$ reward (i.e. punishment) to every step, to encourage more direct game playing. We also use bandit feedback to encourage agents to select less-frequently-used actions \cite{yin2019learn} to avoid repetition. Also, if the game stops early because of a loss, we set the instant reward to $-1$ to penalize the last action.

\subsection{Out-of-genre Games}

Instead of using games in the same cooking genre as the training games, we also generate random games from the TextWorld framework that are in a different genre (treasure hunting). By evaluating our agents on games that are not only unseen but also in a different domain, we can show that our agent learns robust language features that can transfer beyond the specifics of the training data.
We use the TextWorld v1.1.1 to make new games. 
We generate 208 games in total, with 456 entities that never appear in cooking games. These treasure hunting games require the player to navigate around rooms, find a specific entity that is described in each quest, and take a specified action with the entity, e.g.  recovering a chocolate bar from the chest in a dish-pit and eating the
chocolate bar; or picking up the rectangular passkey and inserting it into the rectangular gate's lock in a different room to unlock it. These tasks have no overlap with any cooking games. We  evaluate our agents with these games directly without any further fine-tuning, in the same way we do for cooking test games.

\begin{figure}
\centering
    \includegraphics[width=0.5\textwidth]{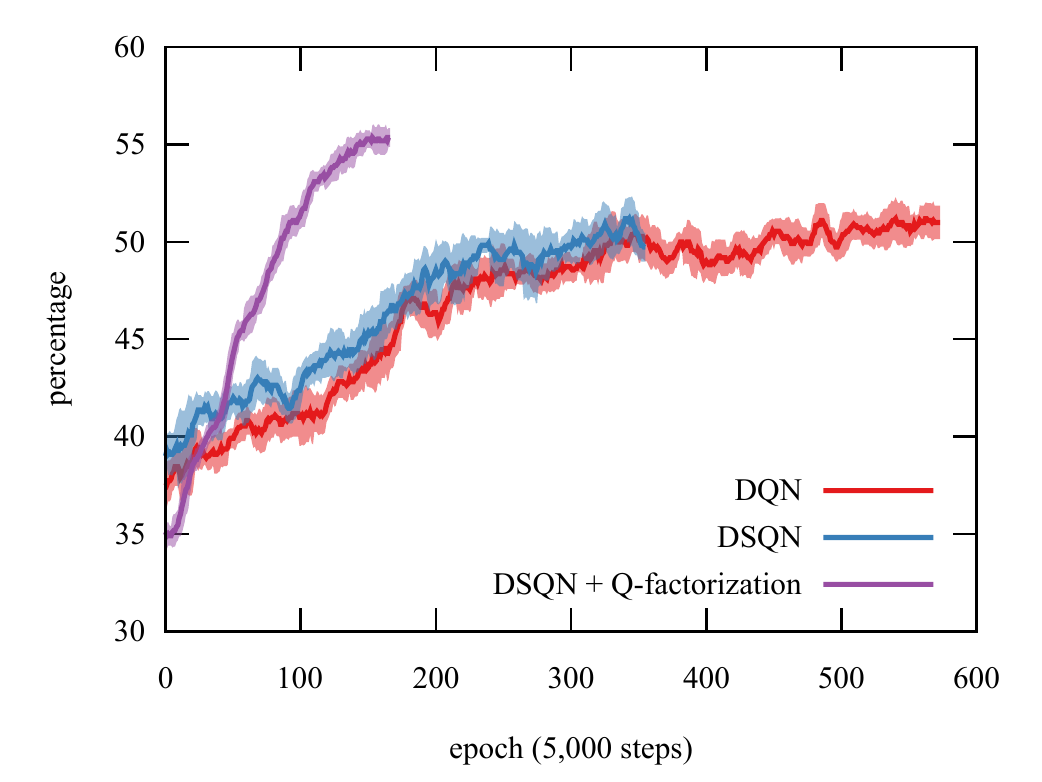}
    \caption{Game-playing scores (percentage of maximum) during training on the dev set, evaluated every 5,000 steps. Three models are evaluated. 1) the normal DQN with no SNN; 2) the DSQN without Q-value factorization; 3) the DSQN with Q-value factorization. Model-2 shows similar quality on dev set with marginal benefits, while model-3 requires fewer epochs and reaches a higher percentage of total score.}
   \label{fig:all-dev-eval}
\end{figure}

\subsection{Encoder Setup}
\label{sec:encoder-setup}
We use Transformer \cite{NIPS2017_7181} as the state encoder. Our Transformer has one layer, the model size is 128, the number of attention heads is 8, and the inner dense layer has 256 dimensions. We use a max-pooling layer over the output of the Transformer as the encoded state. The encoder takes up to 500 tokens as input. We trim trajectory in the head if the length exceeds the limit.
We use a BPE \cite{Sennrich_2016} tokenizer and vocabulary with 30,522 tokens. The word embedding dimension is 64. 
The action encoder is an LSTM with one layer and 32 units. We use the same vocabulary and word embedding dimension as the state encoder, but different word embeddings. The hidden state from the final word of each action is used as the hidden action state. We assume that all actions have a length of less than 10, which is a large enough number considering in our cooking games most actions have a length from two to six.

During DSQN training, we use 50,000 observation steps, 500,000 replay memory entries, and decay $\epsilon$ from 1 to 0.0001 in 10,000,000 steps for training with all games in the training set. 
We train models with games in three configurations. Our baseline model is a DQN agent using the Transformer model (Section \ref{sec:encoder-setup}). We then try the DSQN without Q-value factorization, 
Finally, we add Q-value factorization to the DSQN (Section \ref{sec:q-factorization}).

\section{Results and Analysis}

%
%

\subsection{DSQN Training Process}

\begin{figure}
\centering
    \includegraphics[width=0.5\textwidth]{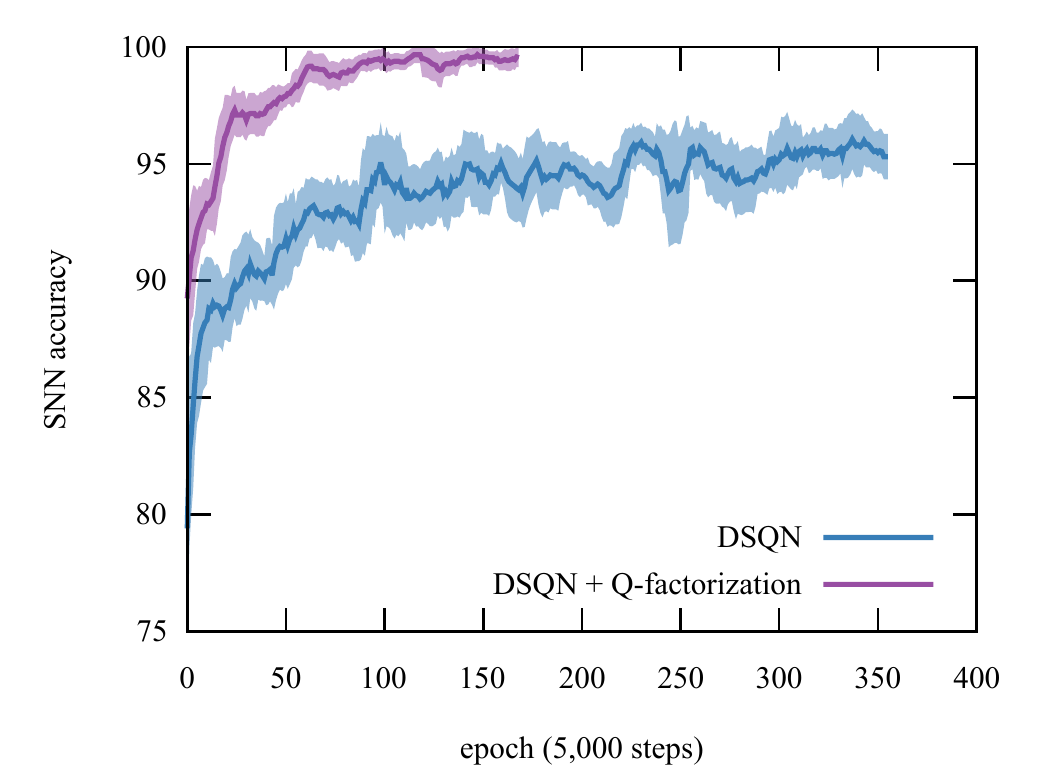}
    \caption{SNN accuracy during training on the dev set (100 pairs of trajectories are extracted at each time). Both curves converge to almost 100 percent accuracy on SNN evaluation much faster than the game-playing score, showing SNN is easier to train than the DQN.}
   \label{fig:all-dev-snn-eval}
\end{figure}


As shown in Figure~\ref{fig:all-dev-eval}, the plain DQN starts from 37\% and ends at 51\%. In a similar number of epochs, the DSQN without using Q-factorization shows that it cannot learn well if there are step-related rewards but is marginally better than the baseline.
The addition of Q-factorization, while starting from a lower point compared to the other two curves, trains much faster and reaches a much higher maximum point, indicating that the Q-value factorization can overcome the problem introduced by step-related rewards.

Figure~\ref{fig:all-dev-snn-eval} shows intrinsic SNN evaluation results during training on the dev set. DSQN with Q-value factorization shows strong SNN accuracy during training by converging to almost 100\% after 100 epochs of training, indicating that the SNN objective is simpler than the full game playing task. Even without Q-factorization, the SNN is able to do a reasonable job at predicting trajectory equivalence. 

\subsection{Zero-shot Game-playing Evaluation}


\begin{table}[]
\centering
\begin{tabular}{lrrr}

dataset  & DQN & DSQN  &  DSQN+fac \\ \hline
test-1    &   53 &   50  & \textbf{57}   \\ 
test-2    &   52 &  50     & \textbf{59}  \\ 
test-th & 24  & 21   & \textbf{42} \\
\end{tabular}
\caption{Extrinsic zero-shot game playing evaluation scores (percentage of total score reached) for five agents. Two same topic (cooking) test sets (test-1 and test-2) and one out-of-genre test set (test-th) are used. DSQN+fac is the DSQN model trained with Q-factorization. DSQN+fac shows better scores than DQN or DSQN without factorization, especially on the out-of-genre treasure hunting games test-th.}
\label{tbl:eval-games-models-games}
\end{table}

From a training run, we select the model with the highest score on the dev set for test inference. We run 10 episodes for each game during the test phase with $\epsilon=0$ and use bandit feedback to dynamically tune Q-values \cite{yin2019learn}. The maximum total score is not unique since different games could have different scores. We use the percentage of scores earned as the evaluation criteria. The higher the score, the better the agent (Table \ref{tbl:clustering-evaluation}). We evaluate three models, 1) the normal DQN, 2) the DSQN (without Q-value factorization), and 3) the DSQN with Q-value factorization. DSQN with factorization shows better scores than the other two models, especially on the out-of-genre treasure hunting games test-th.

We also evaluate the SNN result for DSQN agents, each with 320,000 pairs of trajectories extracted from their evaluation memory (Table~\ref{tbl:eval-games-models-SNN}). The DSQN model with factorization shows high accuracy on all datasets; the DSQN model without factorization only shows good SNN accuracy on test-1 and test-2, the in-genre tests.


\begin{table}[]
\centering
\begin{tabular}{lrr}
dataset      & DSQN      & DSQN+fac   \\ \hline
test-1 & 93.4  & \textbf{99.4} \\ 
test-2 & 92.9  & \textbf{99.4} \\ 
test-th & 69.2  & {\bf 85.0} \\ 
\end{tabular}
\caption{Intrinsic SNN evaluation (state similarity classification accuracy) for DSQN agents. DSQN is trained without Q-value factorization, while DSQN+fac is trained with it. DSQN+fac has higher accuracy on all datasets.}
\label{tbl:eval-games-models-SNN}
\end{table}

\begin{figure*}[ht]
  \centering
  \begin{subfigure}[b]{0.33\linewidth}
    \centering\includegraphics[width=140pt]{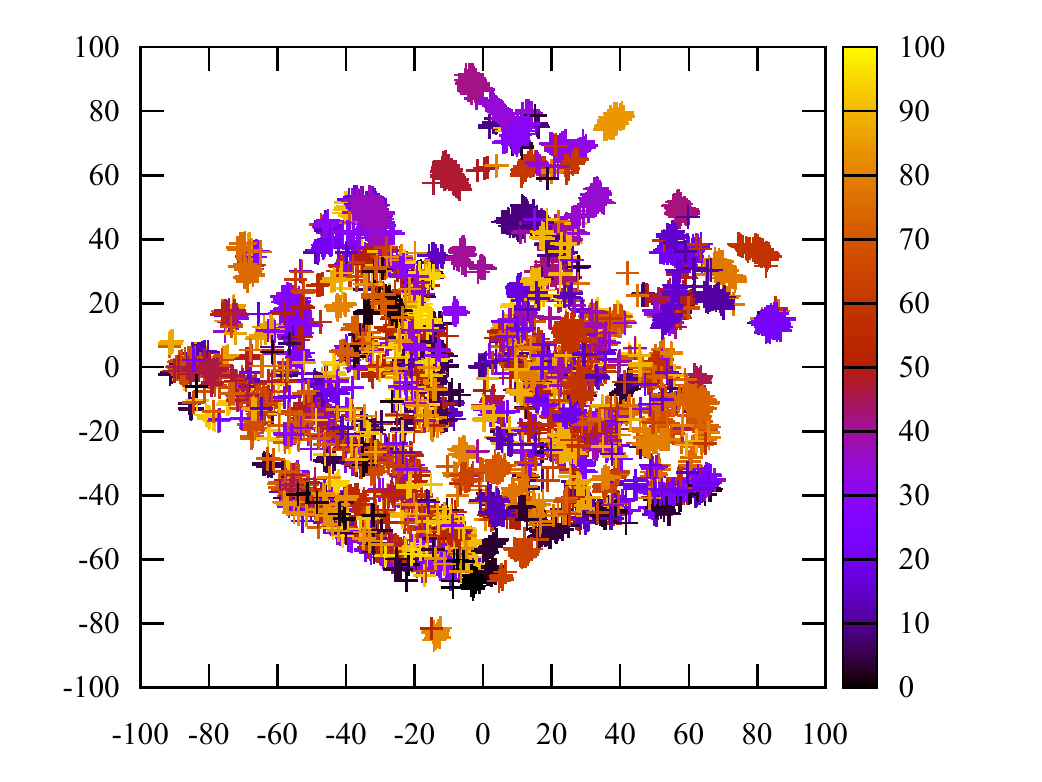}
    \caption{DQN}
  \end{subfigure}%
  \begin{subfigure}[b]{0.33\linewidth}
    \centering\includegraphics[width=140pt]{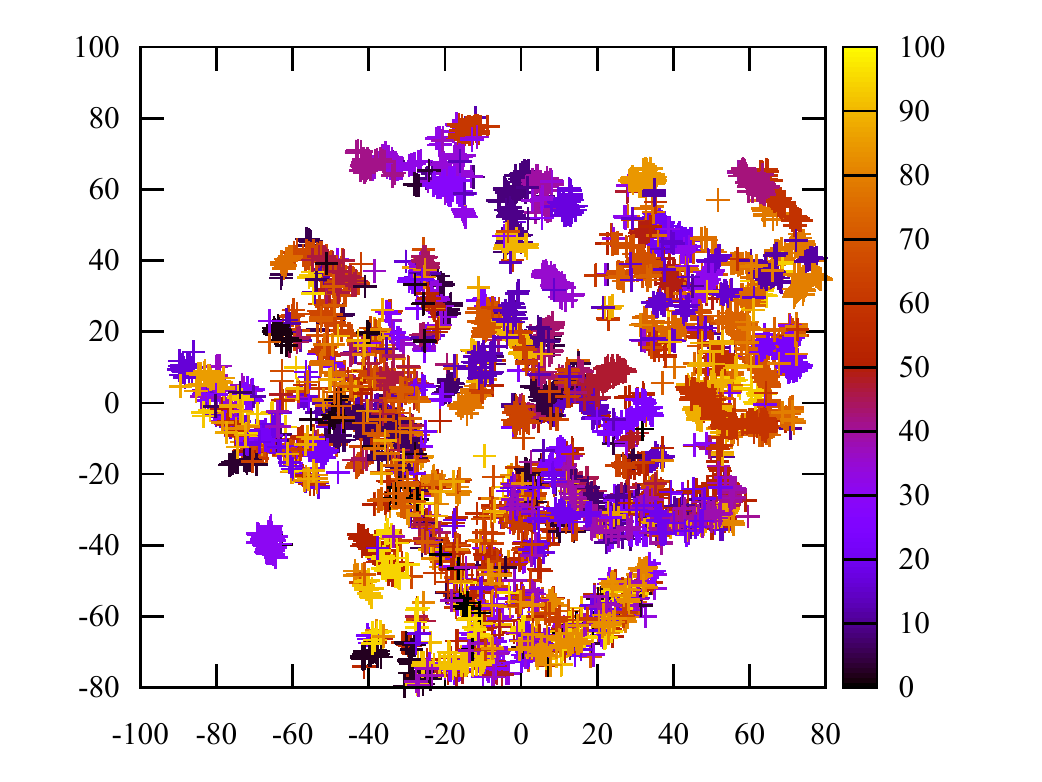}
    \caption{DSQN}
  \end{subfigure}%
   \begin{subfigure}[b]{0.33\linewidth}
    \centering\includegraphics[width=140pt]{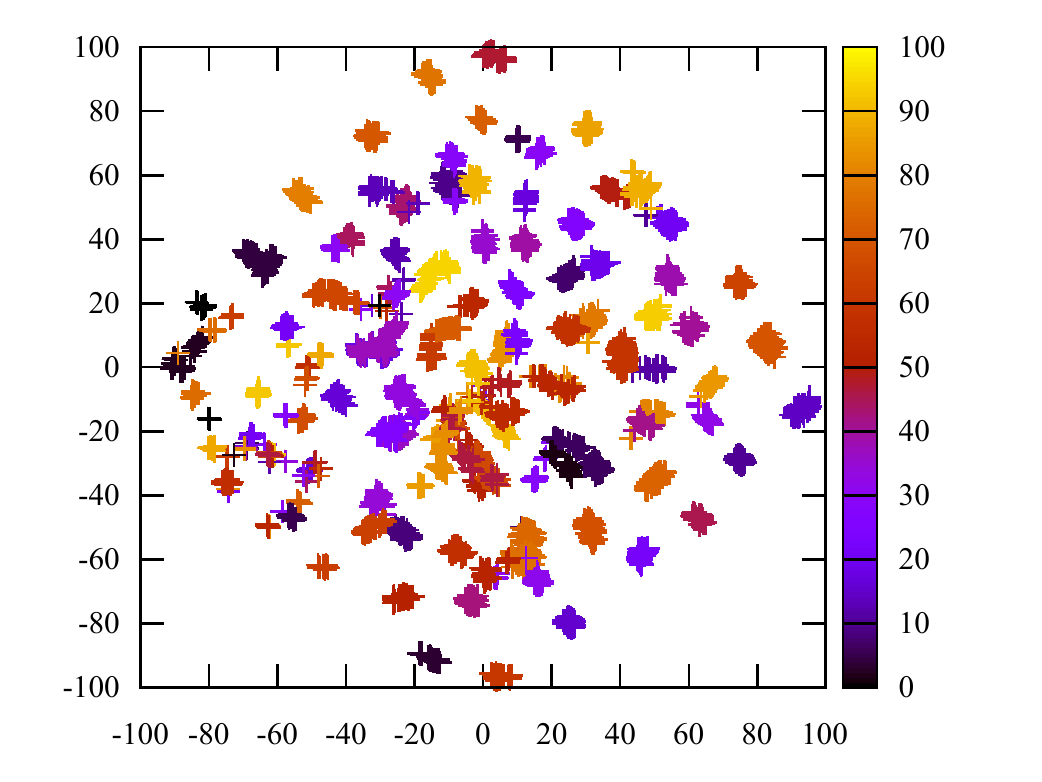}
    \caption{DSQN + Q-factorization}
  \end{subfigure}
  \caption{t-SNE of language features learned by DQN, DSQN, and DSQN  with Q-factorization. There are 95 unique game states and 5,824 trajectories. Different colors indicate different unique game states. DSQN with Q-factorization clusters the game state nicely.}
  \label{fig:tsne}
\end{figure*}

\subsection{Clustering Evaluation}

\begin{table}[]
\centering
\begin{tabular}{lrrr}

model  & H   & C   & V   \\ \hline
DQN    &     .52   &  .61   &  .56      \\ 
DSQN  &   .52  & .60    & .55    \\ 
DSQN+fac    &  \textbf{.87}      &  \textbf{.89}   &  \textbf{.88}      \\ 
\end{tabular}
\caption{Intrinsic clustering evaluation for three agents with H: homogeneity, C: completeness, V: V-measure. The DSQN+fac model generates good language features for clustering.}
\label{tbl:clustering-evaluation}
\end{table}

Since DQN models cannot be evaluated for their SNN classification ability, we use the clustering method to compare the learned features among all learned models. The evaluation metric is V-measure \cite{rosenberg-hirschberg-2007-v}. The V-measure is a combination of homogeneity score and completeness score of clusters, and computationally similar to the F-score. Homogeneity means each cluster should contain only one-class of datapoints, and completeness means each cluster should contain all of these one-class datapoints. The V-measure is weighted over these two measures. These measures give scores between 0 and 1, higher means better clusters.

We run our baseline agents with $\epsilon=0.5$ to allow stochasticity for generating varied trajectories, over 4000 games and generate 500,000 trajectories. We extract 5,824 trajectories from them by selecting labels that appear equal to or more than 50 times. We get 95 unique labels with these trajectories.
Using k-means++ clustering \cite{Arthur:2007:KAC:1283383.1283494}, and $k=95$ (we use scikit-learn; other default hyperparameters are unchanged), we evaluate the clustering labels against our ground truth (hash keys) with the V-measure.

The clustering evaluation result is shown in Table \ref{tbl:clustering-evaluation}. DSQN+fac has a strong clustering evaluation result, with scores around 90\%, while the other models show worse clustering results. Figure \ref{fig:tsne} uses t-SNE \cite{vandermaaten2008visualizing} to extract a visualization of these learned language features. While DQN shows a messy plot with clusters mixed with each other, and DSQN without Q-value factorization also shows an unclear result due to the existence of time-sensitive rewards, the DSQN model with Q-value factorization, clusters equivalent trajectories cleanly.

\section{Related Work}


Other work on text game learning focuses on different approaches to the challenges of DQN with this modality, such as action reduction with language correlation \cite{fulda2017affordance}, a bounding method \cite{DBLP:conf/nips/ZahavyHMMM18}, the introduction of a knowledge graph \cite{DBLP:journals/corr/abs-1812-01628}, text understanding with dependency parsing \cite{Yin2019ComprehensibleCT}, an entity relation graph \cite{DBLP:journals/corr/abs-1812-01628}, and the bandit feedback method for agent evaluation \cite{yin2019learn}. Among these works, state encoding methods vary considerably. \newcite{mnih-atari-2013} use a CNN to understand states from images for video games. \newcite{D15-1001} and \newcite{DBLP:journals/corr/abs-1806-11525} use LSTMs to encode text-based game states, \newcite{DBLP:conf/nips/ZahavyHMMM18} and \newcite{Yin2019ComprehensibleCT} use CNNs  to encode game dialogues with Word2Vec \cite{DBLP:journals/corr/abs-1301-3781} and randomly initialized word embeddings, respectively. \newcite{he-etal-2017-learning} use multilayer neural networks and treat sentences as bag-of-words. \newcite{DBLP:journals/corr/abs-1905-09700} use GloVe \cite{pennington2014glove} as a pre-trained word embedding.

\newcite{yin2018decipherment} use SNNs to learn image features for one-shot alphabets for ciphers. \newcite{Reimers_2019} learn sentence similarity for BERT via SNN training. \newcite{AAAI1816174} use SNNs for music game-playing but without using RL. \newcite{DBLP:journals/corr/GuptaDLAL17} work on transferring knowledge between robots with differently jointed arms that work on the same task by learning invariant feature spaces through contrastive loss and reconstructive loss of different autoencoders.
\newcite{Lample:2017:PFG:3298483.3298548} use a co-training framework to train a DQN and an enemy detection task for video games. However, they can access their main task's ground truth during training 
while we do not assume the existence of such labels.

The original purpose of our work was to find better and more robust state representations than we obtain with normal DQN to improve training speed and zero-shot learning quality for in-domain game playing. However, we also find that our trained agent can be transferred easily across domains. \newcite{DBLP:journals/corr/abs-1708-00133} focus on building transferable agents for action adventure game playing but convert this to a textual descriptions. They consider that language (textual entity descriptions specifically) can be used as an implicit intermediate channel for transfer policies. They also decompose values as we do, into separate transferable and non-transferable portions. A similar game dynamic decoupling is used by \newcite{NIPS2017_6994} and \newcite{DBLP:journals/corr/abs-1801-02268}. Feature learning for RL transfer is also explored by \newcite{Higgins:2017:DIZ:3305381.3305534} using disentanglement and by \newcite{Ammar:2015:UCT:2886521.2886669} using manifold alignment.
Another approach to transfer learning for RL not explored here is multi-agent learning   \cite{parisotto2015actormimic,DBLP:journals/corr/abs-1809-04474}.

\section{Conclusion}

In a language-oriented adversarial framework, such as that posed by text adventure games, imperfect state representations can hamper learning. We use the SNN as a guide to help learn robust language features in such scenarios by comparing whether two trajectories end up in the same state or not. We create DSQN, a multi-task learning framework that co-trains a DQN and SNN, using information from the former to form training data for the latter, and auto-weighting their respective losses. 
We evaluate our trained agents intrinsically (by clustering and SNN evaluation) and extrinsically (by game-playing), on both same-genre but never-before-seen games and totally out-of-genre games, and find out that, sufficiently equipped with a novel Q-value factorization, the DSQN can provide faster training process, higher game-playing scores/task completion rates, and better language features. This can be broadly applied to other language oriented adversarial frameworks such as negotiation dialogues.

\bibliography{acl2020}
\bibliographystyle{acl_natbib}

\end{document}